\documentclass[]{imag-ms-template}

\usepackage[table]{xcolor}
\usepackage{graphicx}
\usepackage{float}
\usepackage{booktabs}
\usepackage{tabularx}
\usepackage{adjustbox}
\usepackage{placeins}
\usepackage{multirow}

\title{Identifying Neural Signatures from fMRI using Hybrid Principal Components Regression} 

\author{Jared Rieck,$^{1\ast}$ Julia Wrobel,$^{2}$ Joshua Gowin,$^{3}$ Yue Wang,$^{1}$ \\ Martin Paulus,$^{4,5}$  Ryan Peterson$^{1,6}$\\
{\small $^{1}$
Department of Biostatistics and Informatics, University of Colorado Anschutz Medical Campus, Aurora, CO, USA}\\
{\small $^{2}$Department of Biostatistics and Bioinformatics, Emory University, Atlanta, GA, USA}\\
{\small $^{3}$Department of Radiology, University of Colorado Anschutz
Medical Campus, Aurora, CO, USA}\\
{\small $^{4}$Laureate Institute for Brain Research, Tulsa, OK, USA}\\
{\small $^{5}$Department of Community Medicine, University of Tulsa, Tulsa, OK, USA}\\
{\small $^{6}$Department of Internal Medicine, Carver College of Medicine, University of Iowa, Iowa City, IA, USA}\\
{\small $^\ast$Correspondence:  jared.rieck@cuanschutz.edu}}

\begin{document} 

\maketitle 


\begin{abstract}

Recent advances in neuroimaging analysis have enabled accurate decoding of mental state from brain activation patterns during functional magnetic resonance imaging (fMRI) scans. A commonly applied tool for this purpose is principal components regression regularized with the least absolute shrinkage and selection operator (LASSO PCR), a type of multi-voxel pattern analysis (MVPA). This model presumes that all components are equally likely to harbor relevant information, when in fact the task-related signal may be concentrated in specific components. In such cases, the model will fail to select the optimal set of principal components that maximizes the total signal relevant to the cognitive process under study, leading to worse prediction and less accurate identification of the brain regions associated with the task. Here, we present modifications to LASSO PCR that allow for a regularization penalty tied directly to the index of the principal component, reflecting a prior belief that task-relevant signal is more likely to be concentrated in components explaining greater variance. Additionally, we propose a novel hybrid method, Joint Sparsity-Ranked LASSO (JSRL), which integrates component-level and voxel-level activity under an information parity framework and imposes ranked sparsity to guide component selection. We apply the models to brain activation during risk taking, monetary incentive, and emotion regulation tasks. Results demonstrate that incorporating sparsity ranking into LASSO PCR produces models with enhanced classification performance, with JSRL achieving up to 51.7\% improvement in cross-validated deviance $R^2$ and 7.3\% improvement in cross-validated AUC. Furthermore, sparsity-ranked models perform as well as or better than standard LASSO PCR approaches across all classification tasks and allocate predictive weight to brain regions consistent with their established functional roles, offering a robust alternative for MVPA.

\end{abstract}

\section{Introduction}

One goal of brain imaging is to understand brain activation patterns to develop biological markers of psychiatric illness, similar to how an electrocardiogram can inform a cardiologist about heart disease. To this end, the use of machine learning techniques to analyze voxel-level fMRI data, particularly to identify brain regions involved in specific cognitive tasks, has become increasingly common in recent years. This approach, widely known as multi-voxel pattern analysis (MVPA), enables researchers to decode task-related neural patterns from high-dimensional brain data. For example, these techniques have been used to determine how much pain a person is experiencing based on brain activation \citep{Wager2013}, and can be applied as a language decoder that accurately reflects word sequences a person is thinking \citep{Tang_2023}. 

MVPA has also been used to explore numerous mental functions in the context of psychiatric disorders. For example, response to reward in the nucleus accumbens during a monetary incentive delay task has been shown to differentiate individuals with bipolar disorder from healthy controls and, among stimulant users in treatment, can indicate potential to return to stimulant use \citep{johnson2019, mortazavi2024}. Similarly, insula activation levels in response to risky decisions can indicate whether individuals in treatment for methamphetamine or cocaine use disorder return to use these drugs in the year after treatment \citep{paulus2005, gowin2014}. Neural patterns of emotion regulation may also reflect psychiatric morbidity, as individuals with social anxiety disorder show different timing of amygdala activation relative to healthy controls when trying to reappraise negative self-beliefs \citep{goldin2009}. These findings support the premise that neural activation to these mental processes may indicate psychiatric disorders, but there remains a need to better model brain patterns to indicate performance of mental tasks.

Among the variety of machine learning methods applied in MVPA, principal components regression regularized with the least absolute shrinkage and selection operator (LASSO PCR) has emerged as one of the most prevalent \citep{Bedi2015, Wager2013, Rieck2024, Knyazev2024, Coll2022, Speer2023}. This technique involves performing principal component analysis (PCA) on voxel-wise brain activity and using the resulting components as inputs to a LASSO regression model that predicts the cognitive task. The LASSO framework applies a penalty that encourages sparsity by shrinking the coefficients of irrelevant or weakly predictive components to zero. By selecting only a subset of components for prediction, the LASSO PCR model helps to isolate neural patterns that are meaningfully associated with the task.

Importantly, standard LASSO models assume that all predictors have an equal prior probability of contributing to the prediction \citep{tibshirani1996lasso, peterson2019ranked, Peterson2022}. However, the most relevant information for brain activation in a task is often concentrated in the first few components, and developing models that reflect signal asymmetry among components may more accurately capture the relevant human brain activity. In general, there may be groups of candidate predictors where the within-group sparsity is expected to differ. To address this, many variants of LASSO regression have been developed that allow the penalty to vary across covariates. Often, the nature of the modification to the LASSO penalty is context-driven. For example, group-based penalties have been applied in scenarios in which variables belong to a set of unique modalities \citep{Boulesteix2017} or in which polynomial terms and/or interaction effects up to a particular order are under consideration \citep{Peterson2022}. In other scenarios, modifications to the LASSO have been proposed to satisfy important theoretical properties such as being an oracle estimator \citep{Zou_2006}.

In the context of LASSO PCR, each principal component (PC) represents a proportion of the total variance observed in the covariates, and this proportion decreases monotonically with the PC index. That is, the first principal component represents a greater proportion of the variability in the data than the second, with the final PC representing the smallest proportion of variability. It is often recommended to exclude PCs that account for only a small proportion of the total variance, typically by retaining only those PCs needed to surpass a predefined threshold of cumulatively explained variance, such as 90\% \citep{James_Witten_Hastie_Tibshirani_2021}. This practice reflects an implicit preference for upstream components, which capture more variance. While such a heuristic is usually reasonable, the choice of threshold is typically arbitrarily selected without any principled justification. In this work, we seek to formalize and improve upon this informal guideline by introducing a tunable weighting system for all of the PCs. Although it is most likely that relevant brain activity is contained within earlier components, it is possible that certain tasks could be encoded within PCs which represent small amounts of variability. For example, if an fMRI task elicits a subtle response amid high amounts of noise, then the majority of the task-relevant signals might be captured primarily by lower-variance PCs, which could be disregarded under a hard-thresholding approach. If the primary variability in the data is neural activity relevant to the cognitive process of interest, we would conversely anticipate that the higher-variance principal components would be more likely to provide signal relevant to the cognitive process. These scenarios are more appropriately handled by incorporating different sparsity expectations for components systematically rather than assuming that sparsity is invariant to the variance partitioned in a PC. Finally, it is notable that principal components provide a limited perspective of the data because they are non-sparse representations of whole-brain activity. Therefore, it may be advantageous to allow models to lend additional predictive weight to individual voxels or groups of voxels in cases where producing such predictive weight via the PCs alone introduces additional noise. 

In this paper, we suggest a ranked-sparsity-based solution which treats sparsity as a monotonic function of the PC index, reflecting the fact that task-relevant signal may be more concentrated in either upstream or downstream PCs, depending on the nature of the cognitive task and data acquisition. Specifically, we propose a modification to LASSO that assigns penalty weights based on the index of the principal components. This allows the model to incorporate non-uniform prior information with respect to which PCs are more likely to capture task-relevant fMRI signal. We further present models that jointly consider principal components and individual voxel-level activity with an accompanying information-parity-based framework for tuning penalty weights. These methods are applied and evaluated in datasets from three task-based fMRI studies, and their performance is compared to a set of competing methods including LASSO principal components regression (PCR) models which apply a uniform penalty to all PCs. 

This paper begins by introducing the traditional LASSO PCR framework and noting how this framework can be modified to introduce sparsity ranking, which we refer to as sparsity-ranked LASSO PCR (SRL). We then introduce the joint sparsity-ranked LASSO (JSRL), which combines sparsity-ranked principal component data with voxel-level activation, and show how parameters can be chosen to achieve prior information parity between principal components and individual voxels. We proceed to compare these sparsity-ranked models with other penalized regression approaches including LASSO PCR by evaluating their performance in classifying fMRI tasks across three distinct fMRI datasets. The brain regions which drive prediction across a selection of the evaluated methods are summarized and explained through visualization of predictive weights in the brain-space.

\section{Methods}

\subsection{Sparsity-Ranked Models}

The LASSO is a regularization technique for linear models that encourages sparsity in the coefficient estimates by introducing a penalty on their absolute magnitude. In the context of discriminating between two fMRI tasks, the LASSO objective function takes the form:

$$-\sum_{i=1}^n\left[y_ilog(p_i) + (1-y_i)log(1-p_i)\right] + \lambda\sum_{j=1}^p|\theta_j|$$

where $y_i$ is the observed binary outcome for a particular subject (in this case, which cognitive task they performed), $p_i$ is the probability of membership to the positive class, $\theta_j$ is the logistic regression coefficient for the $j$-th of $p$ features, and $\lambda$ is a tuning parameter controlling the strength of the $L_1$ penalty. Extensions of the LASSO objective introduce group-specific penalty weights to allow differential regularization across covariate groups. For instance, \citet{Peterson2022} proposed a sparsity-ranking frameworks with the following modified objective:

$$-\sum_{i=1}^n\left[y_ilog(p_i) + (1-y_i)log(1-p_i)\right]+ \lambda\sum_{g=1}^G\sum_{j=1}^{p_g}w_g|\theta_j^g|$$

where $g = 1, \dots, G$ is an index for each of $G$ covariate groups, $p_g$ is the number of covariates in group $g$, and $w_g$ is group-specific penalty weight, often defined as a function of $p_g$. This approach is particularly useful when covariates can be meaningfully partitioned into $L$ groups—for example, separating main effects from polynomial or interaction terms. 

This framework can also be adapted to principal component regression, allowing for component-specific $L_1$ penalties. Principal component analysis (PCA) is a linear transformation that expresses the voxel-level data matrix $X \in \mathbb{R}^{n \times V}$ in terms of orthogonal components ordered by explained variance. In this context, instead of partitioning the covariates into $G$ groups each with dimension $p_g$, we instead let $p_g =1$ and introduce a weighting function $W_k$ defined uniquely for each PC index $k = 1, \dots, K$. The modified objective becomes:

$$-\sum_{i=1}^n\left[y_ilog(p_i) + (1-y_i)log(1-p_i)\right]+ \lambda\sum_{k=1}^KW_k|\beta_k|$$

where $\beta_k$ is the logistic regression coefficient associated with the $k$-th principal component. The model-based probability for subject $i$ is $p_i = \frac{1}{1+exp\left(-\sum_{k=1}^K\beta_kz_{ik}\right)}$, with $z_{ik}$ denoting the score of subject $i$ on principal component $k$. The index $k$ thus reflects both the rank order of the components and the proportion of variance each explains. The form of $W_k$ therefore encodes a prior belief about the relationship between component variance and task-relevant signal. For example, down-weighting components associated with smaller eigenvalues reflects the assumption that low-variance components are less likely to carry predictive signal. In order to capture these relationships, we define the penalty weighting function:

\begin{equation}
W_k =  \begin{cases} 
k^\gamma & \gamma\geq 0 \\
      (K+1-k)^{-\gamma} & \gamma < 0
      \end{cases}
\label{eq:w_ksrl}
\end{equation}

where $k$ is the PC index and $\gamma$ is a tuning parameter determining the shape of the weighting function. 

When $\gamma = 0$, all components receive equal weights, corresponding to the standard LASSO penalty. When $\gamma < 0$, the function applies larger penalties to components with lower indices (i.e., those associated with higher eigenvalues), resulting in a front-weighted penalty. Conversely, when $\gamma > 0$, higher-index components (associated with less explained variance) receive stronger penalties, producing a back-weighted penalty structure. An illustration of how varying $\gamma$ affects the weighting function is shown in the right panel of Figure 1. This weighting function was chosen for both its theoretical flexibility and its desirable symmetry properties: the penalty profiles for $\gamma = -1$ and $\gamma = 1$ are mirror images of one another, thereby applying equal but opposite emphasis to early versus late principal components.

\subsection{JSRL: Jointly Modeling Principal Components and Individual Voxels}

In some cases, a small number of individual voxels may carry highly discriminative information for distinguishing between fMRI tasks. Models that rely solely on principal components as predictors may obscure these effects, since emphasizing a single voxel inherently alters the contribution of other voxels within the same component. To address this limitation, we introduce a hybrid modeling framework in which the design matrix includes both $K$ principal components and $V$ individual voxel-level predictors.

Explicitly, the JSRL model augments the sparsity-ranked PCR framework by including both PCs and voxels as predictors in a penalized logistic regression objective of the form:

\begin{equation}
-\sum_{i=1}^n \Big[y_i \log(p_i) + (1-y_i)\log(1-p_i)\Big] 
+ \lambda \left( \sum_{k=1}^K W_k |\beta_{PC,k}| + \sum_{v=1}^V W_v |\beta_{vox,v}| \right),
\label{eq:jsrl_obj}
\end{equation}

where $y_i$ is the observed binary outcome for subject $i$, $p_i$ is the fitted probability of membership to the positive class, $\beta_{PC,k}$ are coefficients corresponding to the $k$-th PC, and $\beta_{vox,v}$ are coefficients corresponding to the $v$-th voxel. The weights $W_k$ for the principal components are defined as in Equation~\ref{eq:w_ksrl}, allowing different shapes of the penalty function within the PC indices. For voxels, a constant penalty $W_v$ is applied, reflecting the prior belief that each voxel is equally likely to contribute task-relevant signal. Note that the predictive contribution of a given voxel has two components. First, the voxel contributes indirectly through the PCs, via the weighted sum of eigenvector loadings multiplied by their corresponding coefficients $\beta_{PC,k}$. Second, it contributes directly through its own voxel-wise coefficient $\beta_{vox,v}$. The total effect is the sum of these two components.

As is common practice in penalized regression frameworks such as the LASSO, penalty weights are rescaled prior to model fitting. For example, the \texttt{glmnet} implementation of LASSO in R standardizes penalty weights so that they sum to the total number of variables \citep{glmnet}. This normalization facilitates interpretability and stabilizes tuning, but it does not make the overall level of penalization invariant across different weight functions. In the PC-only setting, the relative values of the penalty weights govern model behavior, and multiplying the entire weighting function $W_k$ by a constant has no effect, since the tuning parameter $\lambda$ absorbs this scaling. 

However, when principal components and voxels are modeled jointly as in JSRL, the relative scale of $W_k$ and $W_v$ becomes important. Additionally, the total number of voxels must be considered. As $V$ increases, stronger regularization should be applied to voxel-level coefficients to avoid overfitting and to reflect the prior belief that a voxel representing a smaller population of neurons is less likely to carry meaningful signal. With these considerations in mind, let $j=1, \dots, K+V$ index all predictors in the concatenated design matrix, with the first $K$ entries corresponding to PCs (indexed by $k=1, \dots ,K$) and the remaining $V$ entries corresponding to voxels (indexed by $v=1, \dots, V$). The weighting function can then be written as

\begin{equation}
W_j =
\begin{cases}
  k^{\gamma},            & \gamma \ge 0,\;\; j=k \in \{1,\dots,K\}, \\[4pt]
  (K+1-k)^{-\gamma},     & \gamma < 0,\;\; j=k \in \{1,\dots,K\}, \\[4pt]
  V^{r},               & j=K+v,\;\; v \in \{1,\dots,V\}.
\end{cases}
\label{eq:w_kjsrl}
\end{equation}

where $r$ is a rescaling factor that controls the relative penalty applied to voxels. The choice of $r$ can be tuned via cross-validation over a grid of candidate values. To determine an appropriate search space, we introduce an information parity framework, selecting $r$ such that PCs and voxels contribute equal prior Fisher information.

\subsubsection{Parameter Estimation under Prior Information Parity}

To determine the value of $r$ that achieves prior information parity between the PC and voxel data, we must calculate the prior information contribution of each feature set. Under standard LASSO we impose $\theta_j \sim Laplace(0, 1/\lambda)$ such that the joint prior distribution has the density

$$\pi(\beta) = \prod_{j=1}^p\frac{\lambda}{2}e^{-\lambda|\theta_j|}$$

Under a generalization of the LASSO framework to allow each covariate group to have a unique penalty factor, we may rewrite this joint density as

$$\pi(\beta) \propto \prod_{g=1}^G\prod_{j=1}^{p_g}{\lambda_g}e^{-\lambda_g|\theta_j^g|}$$

Where $p_g$ is the number of features in the data modality $g$. Here, we have two groups and apply the penalty $\lambda_k = \lambda W_j$ for each PC index $k$ and voxel index $v$. First, let us consider the case that $\gamma \geq 0$. We may then substitute $\lambda^*_k = \lambda k^\gamma$ and $\lambda^*_v = \lambda V^r$ and the joint prior becomes

$$\pi(\beta) \propto \prod_{k=1}^K \lambda^*_ke^{-\lambda^*_k|\beta_k|}\prod_{v=1}^{V}\lambda^*_ve^{-\lambda^*_v|\beta_v|}$$

With corresponding log-density

$$log(\pi(\beta)) \propto \sum_{k=1}^K log(\lambda^*_k)-\lambda^*_k|\beta_k| + \sum_{v=1}^{V}log(\lambda^*_v) - \lambda^*_v|\beta_v|$$

The Fisher information with respect to the parameter $\lambda^*_k$ for a single PC is then

$$I(\lambda^*_k) = \frac{1}{\left(\lambda^*_k\right)^2} = \frac{1}{\lambda^2k^{2\gamma}}$$

The total contribution of all $K$ PC covariates to the Fisher information is then

$$I_{total}^K = \sum_{k=1}^K I(\lambda^*_k) = \sum_{k=1}^K \frac{1}{\left(\lambda^*_k\right)^2} = \frac{1}{\lambda^2}\sum_{k=1}^K \frac{1}{k^{2\gamma}} = \frac{1}{\lambda^2}H_{K, 2\gamma}$$

Where $H_{k, 2 \gamma}$ is the $k^{th}$ harmonic number of order $2\gamma$. Similarly, the Fisher information with respect to $\lambda^*_v$ for one voxel is

$$I(\lambda^*_v) = \frac{1}{\left(\lambda^*_v\right)^2}$$

With corresponding information for all voxels

$$I_{total}^V = \sum_{v=1}^V I(\lambda^*_v) = \sum_{v=1}^V \frac{1}{\left(\lambda^*_v\right)^2} = \sum_{v=1}^V \frac{1}{\left(\lambda V^r\right)^2} = \frac{V}{\left(\lambda V^r\right)^2} = \frac{1}{\lambda^2 V^{2r-1}}$$

Under this formulation, we achieve information parity when

$$\frac{1}{\lambda^2}H_{K, 2\gamma}=\frac{1}{\lambda^2 V^{2r-1}} \implies H_{K,2\gamma} = V^{1-2r}$$

Solving for $r$ yields

$$log\left(H_{K, 2\gamma}\right) = (1-2r)log(V) \implies r=\frac{-log\left(H_{K,2\gamma}\right)}{2log(V)}+\frac{1}{2}$$

Therefore, when $\gamma \geq 0$ the rescaling factor can be chosen to produce a penalty for voxel indeces that produces information parity between the PC and voxel data modalities when

$$r_{IP}=\frac{-log\left(H_{K,2\gamma}\right)}{2log(V)}+\frac{1}{2}$$

The same rescaling factor can be chosen when $\gamma <0$, but we substitute $\gamma = |\gamma|$, yielding the same total information contribution of PCs. This leads to a final expression for the rescaling factor 

\begin{equation}
r_{IP}=\frac{-log\left(\sum_{k=1}^K\frac{1}{k^{2|\gamma|}}\right)}{2log(V)}+\frac{1}{2}
\label{eq:r_value}
\end{equation}

In practice, the value $r_{IP}$ which achieves prior information parity may not produce the best-performing model, as it may be that there is more or less signal in the individual voxels than is reflected by this penalty. Therefore, the information parity value $r_{IP}$ is used as a highly interpretable starting point for cross-validation, and various penalty multipliers ($\tau$) are considered to allow for more or less penalization of the voxel modality than is reflected by this value. 

\subsection{Datasets}

We test our methods on data from three separate task-based fMRI studies: Emotion Regulation (ER; \cite{Rieck2024}), Monetary Incentive Delay (MID; \cite{Kirk‐Provencher_Hakimi_Andereas_Penner_Gowin_2024}), and Risky Gains (RG; \cite{gowin2014}). The Emotion Regulation and Monetary Incentive Delay studies shared a cohort of eighty-two subjects (mean age=20.95 years, SD=1.34, 54.2\% Female, 77.2\% Caucasian) recruited from the Denver metro area. 

In the ER dataset, subjects were shown images from the International Affective Picture System (IAPS) which is widely used in psychological to study emotion and attention \citep{Bradley_Lang_2017}. Participants viewed 15 neutral and 30 negative images. Neutral images were always paired with a 'look' instruction, directing participants to simply observe the image (Neutral class). Of the negative images, 15 were also paired with 'look' (Negative class), while the remaining 15 were paired with a 'decrease' instruction, prompting participants to regulate their emotional response and feel less negatively (Decrease class).

In the MID dataset, participants completed trials in which they anticipated and responded to cues signaling potential monetary outcomes: gaining \$5 (Win5), losing \$5 (Lose5), or no change in earnings (Win0). Trials were preceded by a cue and condition indicator—safe (yellow border) or threat (blue border), where threat signified the potential for an aversive stimulus (scream and scary face). Participants had to respond quickly to a brief target to win or avoid losing money, depending on the cue type. Trials were evenly distributed across six cue-condition combinations (e.g., Win5–threat, Lose5–safe), with 18 trials per type across two runs. Data were categorized by trial outcome—Win5, Win0, and Lose5—for modeling purposes. The rare presentation of the actual scream was excluded from analysis to isolate the anticipatory effect of threat.

The Risky Gains study consists of 68 methamphetamine-dependent individuals (mean age=38.2 years, SD=10.5, 22.1\% Female, 60.3\% Caucasian) recruited through 28-day in-patient alcohol and drug treatment programs (ADTP) at the San Diego Veterans Affairs Medical Center and Scripps Green Hospital (La Jolla, CA,USA). All subjects ceased using methamphetamine for an average of 34.0 ± 3.4 days prior to participation (range 15–207) and were randomly screened for the presence of drugs as part of the treatment program. In the Risky Gains Task, participants completed 96 trials where they earned points to be exchanged for money by choosing one of three ascending options (20, 40, or 80 points) presented sequentially. Choosing 20 (the safe option) always awarded points. Choosing 40 or 80 (the risky options) could result in either a reward or an immediate loss ($-$40 or $-$80). Unbeknownst to participants, all strategies led to the same final point total, controlling for reward bias. Trials were categorized into three conditions based on decision type and timing: baseline activation (pre-decision), safe decisions (20-point choices), and risky decisions (40- or 80-point choices).

Comprehensive descriptions of sample characteristics, inclusion/exclusion criteria, designs, procedures, scanning parameters, and preprocessing for all three studies can be found in prior publications \citep{gowin2014, Rieck2024, Kirk‐Provencher_Hakimi_Andereas_Penner_Gowin_2024}.

\subsection{fMRI Data Preprocessing}

Data preprocessing steps were performed using Analysis of Functional NeuroImages (AFNI) Version 22.0.2141 \citep{cox1996afni}. Each image was converted into AFNI-compatible formats and echoplanar images were aligned with anatomical images. Spatial normalization was performed via nonlinearly warping of anatomical data to the Montreal Neurological Institute (MNI) standard space using AFNI SSwarper42. Images were skull-stripped and deobliqued using the MNI152 template. Time points with more than 0.3-mm Euclidean distance of framewise displacement were censored from analyses. An 8-mm kernel was used for blur for emotion regulation and monetary incentive delay datasets, while a 4-mm kernel was used for the Risky Gains dataset. Each run was scaled to produce a mean intensity for each voxel of 100. A 4-s block model was applied to regression analysis to obtain voxel-level beta weights for each class of stimulus. Regressors of interest were the presentation of each cue for the corresponding task (e.g. decrease-negative, safe decision, etc.). For the monetary incentive delay and emotion regulation datasets, six regressors of no interest were included to account for motion of translation and rotation in the x, y, and z dimensions. For the risky gains dataset, three motion regressors (roll, pitch, and yaw) were included alongside a regressor for slow linear drift as regressors of no interest.

\subsection{Model Comparison}

\begin{figure}[H]\begin{center}\includegraphics[width=0.8\textwidth]{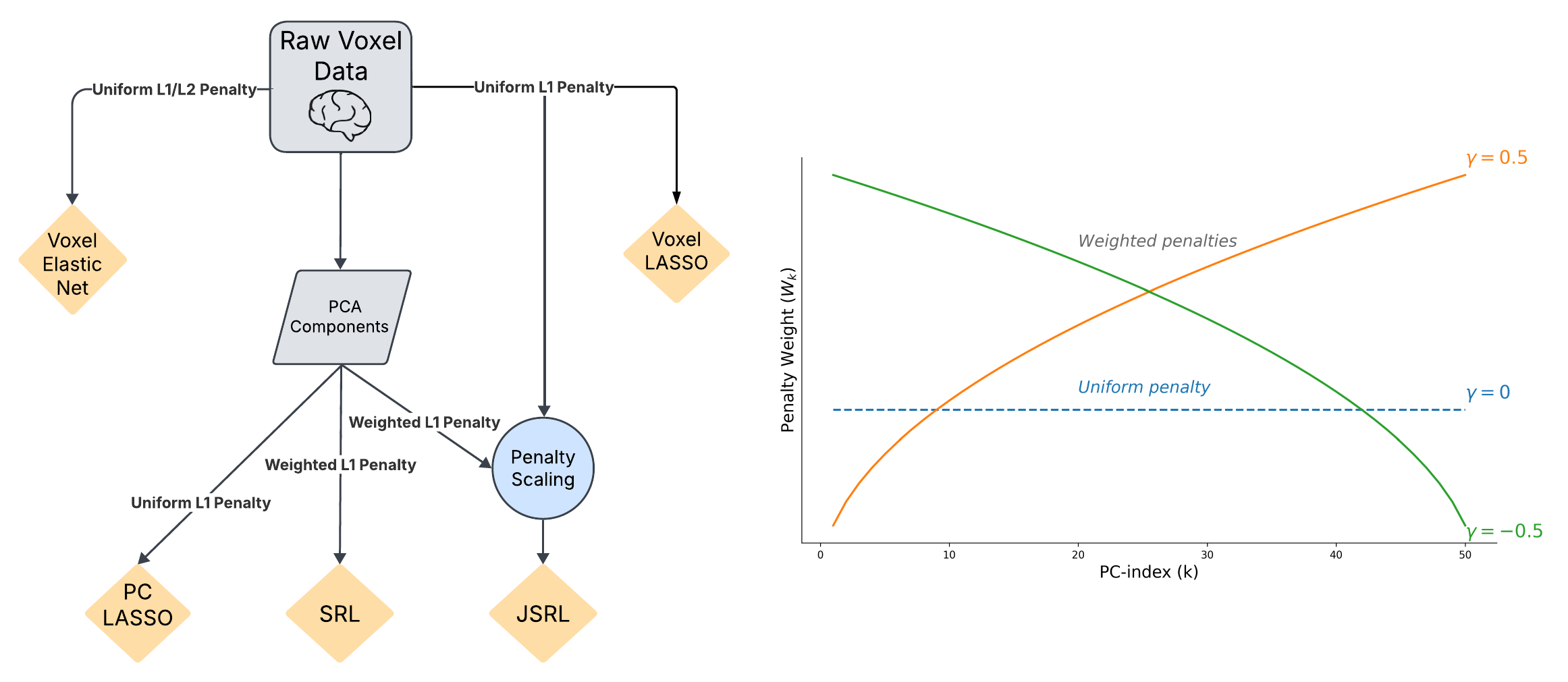}
\caption{Conceptual framework of modeling methods and sparsity ranking for principal components. A uniform penalty ($\gamma = 0$) is displayed alongside two ranked penalty functions, one for a back-weighted penalty ranking ($\gamma >0$) and one for a front-weighted penalty ranking ($\gamma <0)$.}
\label{fig1}\end{center}\end{figure}

A total of five modeling methods were applied to single-subject contrast maps to produce predictive models for each of the three task-based fMRI studies. For each method, performance metrics were averaged over 10 runs, with each run using a different random seed to determine training/testing splits during 10-fold cross-validation. This approach ensured robust tuning parameter estimation and fair performance comparison. The construction of these six models is schematically illustrated in the left panel of Figure 1. All models were fit using the glmnet R package \citep{glmnet}. Prior to modeling, voxel-level beta weights were standardized to have zero mean and unit variance. Within glmnet, predictors are further standardized internally to have mean 0 and unit $L_2$ norm before penalization, ensuring that penalties are applied uniformly across predictors.

Sparsity-ranked LASSO (SRL) uses PCs as predictors and applies penalty weights as in Equation~\eqref{eq:w_ksrl}. Cross-validation was performed to select the $\gamma$ parameter for SRL models using a set of 27 candidate $\gamma$ values ranging from -3 to 3 with a denser grid about $\gamma=0$. For each $\gamma$, the penalty weight $\lambda$ was selected via 10-fold cross validation. Joint sparsity-ranked LASSO (JSRL) uses PCs and individual voxels as predictors and applies penalty weights as in Equation~\eqref{eq:w_kjsrl} using the same set of candidate $\gamma$ values as for SRL. To fit this model, a nested cross-validation procedure was employed in which for every candidate $\gamma$ value the corresponding $r_{IP}$ which achieves information parity as per Equation~\eqref{eq:r_value} was used to determine the penalty weights on voxels. For each such $(r_{IP}, \gamma)$ pair, the penalty weight $\lambda$ was then selected via 10-fold cross validation. Then, the $\gamma$ value was extracted from the model which minimized the cross-validated deviance. A final cross-validation step was performed to determine the optimal value for the voxel-mode penalty from a set of candidate $\tau$ values which modify the prior information parity value by a multiplicative factor, i.e. $r = \tau \times r_{IP}$. The values for $\tau$ ranged from 0.1 to 1.5, with the penalty weight $\lambda$ again being selected via 10-fold cross validation for each such model. The final model selected for prediction used the parameter values $r,\gamma, \lambda$ from the model which minimized the cross-validated deviance in this step. 

We compare our proposed SRL and JSRL methods with several alternative approaches including Principal Component LASSO (PCL) which uses PCs as predictors and applies a constant $L_1$ penalty to each covariate as well as voxel-LASSO (VL) and voxel-elastic net (VEN) which use only individual voxels as predictors and apply a constant penalty to each covariate. For the voxel-elastic net model, a mixing parameter of $\alpha=0.5$ was applied.

\subsection{Brain Activation Maps}

A subset of models utilizing information derived from principal components including PCL, SRL, and JSRL were selected for visual comparison using brain activation maps. To construct these visualizations, statistical maps were extracted for each model by projecting the $\beta$ coefficients for each principal component onto the brainspace by multiplying them by the corresponding eigenvector, and then summing the results across all PCs. For JSRL, the total activation for each voxel was computed by additionally adding the $\beta$ coefficients estimated directly via the voxel-wise LASSO component. 

These statistical maps were visualized in the MNI152 template space and thresholded to highlight voxels with higher activation values. Activation maps corresponding to two selected classification tasks (Decrease vs. Negative and Decrease vs. Neutral) were compared across methods. To ensure spatial alignment, a single brain slice was chosen for each task. This reference slice was determined by identifying the slice that maximized the total sum of absolute voxel-wise $\beta$ values in the Principal Component LASSO (PCL) model, and the same reference was applied to all models for that particular task to facilitate visual comparison.

\subsection{Region Annotation and Volume Aggregation}

Individual voxels that receive additional weight through the JSRL represent regions (or neuron clusters) which strongly associate with the task, in particular when JSRL achieves stronger performance than SRL. To highlight the regions which these JSRL weights comprise, the we annotated voxels that received additional predictive weight via the raw voxel data components in Emotion Regulation tasks with anatomical regions from the Harvard–Oxford (HO) atlases and summarized regional volumes \citep{HarvardOxford_RRID_SCR_001476}. Voxel-wise predictive/statistical maps (native 3 mm isotropic) were resampled to the HO 1 mm cortical atlas grid using nearest-neighbor interpolation to preserve discrete indices \citep{Abraham2014_Nilearn}. Each voxel was assigned a single anatomical label by querying the HO cortical max-probability atlas first; if no cortical label was present at that location, the HO subcortical max-probability atlas was queried. Regional volumes were computed as the count of labeled 1 mm voxels, which in this grid is numerically equal to cubic millimeters (mm³). 

\section{Results}

\begin{figure}[H]\begin{center}\includegraphics[width=0.8\textwidth]{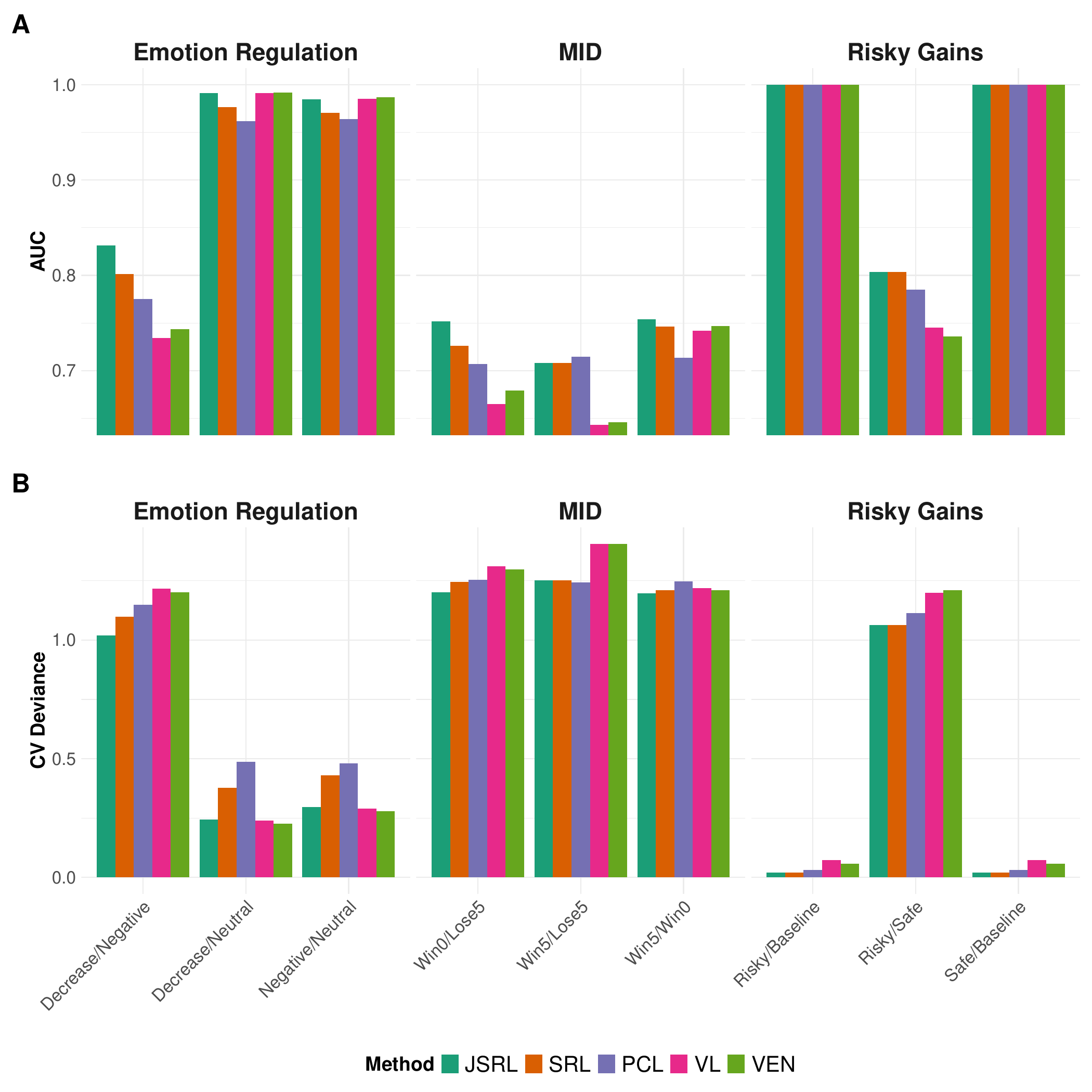}
\caption{Classification performance for each method is compared in terms of cross-validated AUC (A) and average cross-validated binomial deviance (B) achieved by each modeling method for each of 9 total binary classification tasks across the three datasets.}
\label{fig2}\end{center}\end{figure}

JSRL achieves more than a 5\% increase in cross-validated AUC over the standard PCL method in three classification tasks and attains the highest (or joint-highest) AUC in the remaining tasks (Figure 2, Figure S1). It also yields the lowest cross-validated deviance in 6 of 9 tasks and performs comparably to the best method in the remaining three (Figure 2, Figure S2). Cross-validated deviance $R^2$ values for SRL and JSRL are likewise highest in 6 of 9 tasks, with JSRL achieving improvements of up to 51.7\% relative to PCL (Figure S3). Both SRL and JSRL outperform PCL on all emotion-regulation tasks and outperform ordinary LASSO on cross-validated deviance across all risky-gains tasks. For transparency, Figures S1 and S2 also present the metrics as percentage changes relative to PCL

\begin{table}[H]
\centering
\begin{tabular}{llrrr}
\toprule
Dataset & Task & $\gamma$ & $r_{IP}$ & $\tau$\\
\midrule
Emotion Regulation & Decrease/Negative & 0.300 & 0.197 & 0.395\\
Emotion Regulation & Decrease/Neutral & 0.325 & 0.057 & 0.115\\
Emotion Regulation & Negative/Neutral & 0.385 & 0.065 & 0.130\\
MID & Win5/Lose5 & 0.030 & 0.412 & 0.825\\
MID & Win5/Win0 & 0.250 & 0.205 & 0.410\\
\addlinespace
MID & Win0/Lose5 & 0.255 & 0.230 & 0.460\\
Risky Gains & Risky/Baseline & 1.000 & 0.324 & 0.650\\
Risky Gains & Safe/Baseline & 1.000 & 0.324 & 0.650\\
Risky Gains & Risky/Safe & 0.270 & 0.449 & 0.900\\
\bottomrule
\end{tabular}
\caption{Average JSRL Parameters by Classification Task. The multiplier column reports the factor by which the information parity r-value is scaled in the JSRL model.}
\end{table}

Table 1 summarizes, for each classification task and across 10 random-seed repetitions, the average values of the sparsity‐ranking exponent $\gamma$, the information-parity value $r_{IP}$, and the multiplicative rescaling factor $\tau$ applied to $r_{IP}$ for JSRL models. Positive $\gamma$ values correspond to down-weighting of later PCs (i.e. reduced penalization of early components), whereas negative $\gamma$ values correspond to up-weighting of later PCs. Notably, the Win5/Lose5 models consistently select near-zero $\gamma$, indicating no meaningful benefit from sparsity ranking. By contrast, all other tasks select nonzero $\gamma$, reflecting front-loading of the most informative PCs.

The Risky Gains Risky/Baseline and Safe/Baseline tasks show the largest $\gamma$ values both averaging $\gamma=1$, indicating the strongest relative penalization of downstream PCs. This suggests that the cognitive processes these tasks engage are primarily captured by the first few PCs, which explain the greatest variance in the data. Correspondingly, all models are able to achieve perfect classification performance in terms of AUC, as careful variable selection is not needed to produce a model that separates these classes.

Larger values for $\tau$ indicate larger penalties to the voxels than is reflected under information parity, rendering it more difficult for voxels to be selected relative to PCs. $\tau$ values were all \textless $1$, indicating that for these datasets, models could be improved by expecting more signal to come from PCs than voxels. The emotion regulation dataset comprised the smallest $\tau$ values ranging from 0.115 to 0.395, indicating high levels of detectable voxel-level activity associated with the classification tasks, especially those that are involved in distinguishing from the Neutral condition.

\begin{figure}[H]\begin{center}\includegraphics[width=0.8\textwidth]{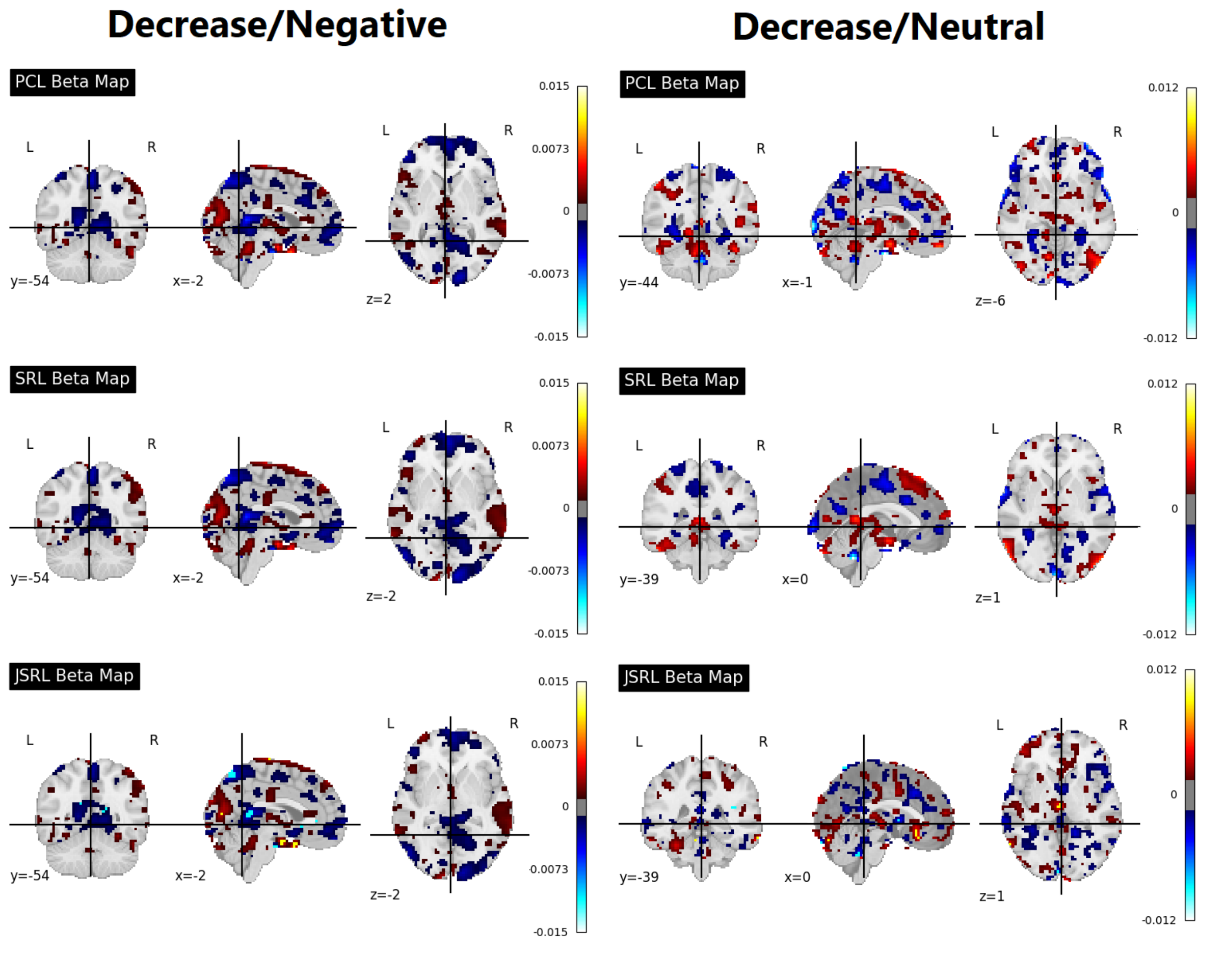}
\caption{Distribution of beta coefficient values in the brain space for PCL, SRL, and JSRL models are displayed for two separate classification tasks in the emotion regulation dataset. For all models, red displays regions that had activity which was predictive of the positive class (Decrease) and blue displays regions that had activity which was predictive of the negative class (Negative or  Neutral). For JSRL models, bright blue displays voxels with activity that was strongly predictive of the negative class, and yellow displays voxels with activity that was strongly predictive of the positive class.}
\label{fig3}\end{center}\end{figure}

Figure 3 illustrates the top-weighted brain regions driving the Decrease/Negative and Decrease/Neutral classification models that use principal-component inputs. Across all modeling approaches, the same anatomical loci emerge as the key predictors, with their relative importance differing only in the spatial extent of the weight clusters and the internal weight distributions. JSRL is distinguished by a tendency to concentrate weight at the very centers of those predictive clusters through the voxel-wise weights, whereas PCL and SRL distribute weight more diffusely across the region.

\begin{table}[htbp]
  \centering
  \caption{Regions with highest volumes of activated individual voxels for Emotion Regulation JSRL models, ranked by descending total volumes.}
  \label{tab:top_regions}
  \begin{tabularx}{\linewidth}{@{}l c >{\raggedright\arraybackslash}X r@{}}
    \toprule
    \textbf{Task} & \textbf{Rank} & \textbf{Region} & \textbf{Total Volume ($\text{mm}^3$)} \\
    \midrule
    \multirow[t]{5}{*}{Decrease vs.\ Negative}
      & 1 & Precuneus Cortex & 155 \\
      & 2 & Subcallosal Cortex & 108 \\
      & 3 & Precentral Gyrus & 100 \\
      & 4 & Superior Frontal Gyrus & 91 \\
      & 5 & Lingual Gyrus & 89 \\
    \addlinespace
    \multirow[t]{5}{*}{Decrease vs.\ Neutral}
      & 1 & Temporal Occipital Fusiform Cortex & 152 \\
      & 2 & Lingual Gyrus & 109 \\
      & 3 & Temporal Pole & 99 \\
      & 4 & Left Cerebral Cortex & 84 \\
      & 5 & Frontal Orbital Cortex & 83 \\
    \addlinespace
    \multirow[t]{5}{*}{Negative vs.\ Neutral}
      & 1 & Lateral Occipital Cortex, inferior division & 234 \\
      & 2 & Frontal Pole & 161 \\
      & 3 & Temporal Occipital Fusiform Cortex & 105 \\
      & 4 & Left Cerebral Cortex & 98 \\
      & 5 & Lingual Gyrus & 92 \\
    \bottomrule
  \end{tabularx}
\end{table}
\FloatBarrier

Across Emotion Regulation JSRL models, predictive voxels localized to a compact set of visual and frontal–limbic territories (Table 2). For Decrease vs. Negative, volume was led by Precuneus Cortex, with additional contributions from subcallosal and superior frontal regions along with Precentral Gyrus. Decrease vs. Neutral emphasized ventral visual areas, with Temporal Occipital Fusiform Cortex and Lingual Gyrus ranked highest. Negative vs. Neutral was dominated by Lateral Occipital Cortex (inferior division) and Frontal Pole, with involvement of fusiform and lingual regions across tasks. Both visual processing (lateral occipital, fusiform, lingual) and control/valuation networks (frontal pole, superior frontal, subcallosal, precuneus) carry substantial predictive weight in the voxel-wise component of JSRL.

\section{Discussion}

We examined whether classification of mental states using fMRI data can be improved by incorporating sparsity ranking. The results demonstrate that sparsity-ranked penalized principal component regression frameworks enhance classification performance as compared to standard LASSO principal components regression models. Both the Sparsity Ranked LASSO (SRL) and Joint Sparsity Ranked LASSO (JSRL) consistently outperformed or matched the performance of the standard Principal Components LASSO (PCL) model, and achieved substantial gains in predictive performance for some classification tasks, with JSRL exhibiting additional gains in tasks that benefit from adding voxel-level predictive weights. This superlative performance suggests that emphasizing sparsity not only at the level of individual principal components but also across sets of voxels with high associations to the cognitive process under study can improve the model's capability to correctly identify in which mental states an individual is engaged during a task using fMRI data.

Activation maps produced by different models were visualized in emotion regulation tasks where sparsity-ranked models exhibited the largest gains in predictive performance. These visualizations show that whereas principal component-only models produce diffuse distribution of weights, JSRL tends to concentrate predictive weight in the centers of clusters of predictive weights. This focal weighting suggests that JSRL preferentially weights the subregions containing the highest density of neurons driving the cognitive process, yielding greater predictive performance when such “core” populations dominate task representations.

These additional voxel-level weights occur within relevant functional regions, enhancing the neuroscientific interpretability of JSRL models. For instance, precuneus cortex represents the largest volume of voxels receiving additional weight in JSRL models distinguishing passive viewing of negative imagery from instructed emotional down‑regulation, matching prior research that demonstrated enhanced connectivity between the precuneus and amygdala when individuals directed attention away from emotional stimuli \citep{ferri2016emotion}. The precentral and frontal gyri are also given additional predictive weight by JSRL, mirroring findings of emotion regulation-relevant regions highlighted in prior fMRI analyses \citep{keller2022transdiagnostic, Rieck2024}. Regions identified for other emotion regulation classification tasks identified large volumes of weights in prefrontal cortical regions including frontal pole and frontal orbital cortex, matching the described importance of these regions in meta analyses studying emotion regulation signatures \citep{picoperez2017emotion, morawetz2017effect, kohn2014neural}. These up-weighted regions are not uniquely identified by JSRL, as there was strong agreement across models regarding which brain regions were most positively associated with task classification, reinforcing the robustness of the discovered patterns. Rather, other regularized PCR models distribute weight more diffusely across an ROI’s extent, potentially diluting the signal from the most informative voxels. Edges of anatomically defined ROIs are particularly susceptible to partial-volume effects, lower signal-to-noise, and inter-subject registration error, which can introduce noise and reduce statistical power \citep{poldrack}. By placing minimal weight on boundary voxels, JSRL effectively “automates” a form of noise filtering, mitigating contamination from neighboring tissue and smoothing artifacts.

Alternative approaches to constructing sparsity-ranked models were considered but ultimately not included in the final analysis. One such approach involved assigning penalty weights directly as a function of the principal component eigenvalues, rather than indirectly linking weights to eigenvalues through the component index. However, this approach yielded nearly identical models to the index-based penalty, with no discernible differences in classification performance. Given this, the index-based formulation was preferred for its parsimony and ease of interpretation. In parallel, we also evaluated the effect of repeated cross-validation on the JSRL framework. This analysis showed no significant improvement in predictive performance. The selected tuning parameters for JSRL remained consistent across cross-validation splits, suggesting that a single round of tuning may be sufficient for reliable model performance, which is an important consideration for computationally intensive neuroimaging studies. Finally, we compared our methods with classical principal component regression approaches, such as keeping a predefined number of components based on a threshold of cumulatively explained variance or determined via cross-validation. Such methods which implemented a hard threshold on the number of components to retain uniformly underperformed the models considered here, and were omitted from results.

Parameter selection for JSRL models performed via cross-validation often selected prior information multipliers ($\tau$) below one, indicating that optimal performance was achieved by imposing weaker penalties on voxel-level coefficients than under prior information parity. This may reflect the advantage of allowing the model to emphasize localized, high-information voxels, especially when predictive signal is concentrated within specific brain regions. Notably, selected values for $\tau$ were generally higher for the Risky Gains classification tasks, likely indicating that the patterns needed to distinguish between tasks are sufficiently represented by the principal components, reducing the benefit of voxel-level contributions. 

Several limitations of this work should be acknowledged and addressed in future research. First, model performance was evaluated solely through cross-validation, without validation in an independent holdout dataset. Future studies should confirm generalizability across broader datasets and task paradigms, and extend the analysis to nonlinear modeling approaches. Second, although the PCA preprocessing step is unsupervised and analogous to centering and scaling, it introduces a potential avenue for data leakage that could influence out-of-sample fits. Finally, our evaluation of interpetability of the voxel-wise component of JSRL models is primarily qualitiative, and could be strengthened through stability selection maps across folds/seeds, which is left to future work. While these limitations warrant caution, our results nonetheless suggest that ranked sparsity provides a promising path toward developing more accurate and interpretable models of brain function.

\section*{Data and Code Availability}

Code will be made available on GitHub upon manuscript acceptance. Data are available from the corresponding author upon reasonable request.

\section*{Author Contributions}

Analyses were performed by J.R. with support from R.P. and J.W. Writing of the main manuscript text and the preparation of all figures was performed by J.R. All authors discussed the results and contributed to the final manuscript. J.R. will serve as the corresponding author.

\section*{Declaration of Competing Interests}

The authors declare no competing interests.

\section*{Supplementary Material}

This supplement contains three figures that parallel Figure 2, but report results relative to principal component LASSO (PCL), the reference model. While raw AUC and deviance provide absolute performance measures, deviance is reported on a log-likelihood scale that is not always intuitive. To aid interpretability, we present values expressed as differences from PCL and further compute cross-validated deviance $R^2$, which rescales out-of-sample deviance relative to the baseline. This unitless metric parallels the familiar $R^2$ in regression, quantifying the proportion of deviance explained beyond PCL. These figures therefore illustrate both the magnitude of improvement achieved by sparsity-ranked models (SRL, JSRL) and highlight the differential classification performance of voxel-level models (VL, VEN) across tasks.

\renewcommand{\thefigure}{S\arabic{figure}}
\setcounter{figure}{0} 
\begin{figure}[H]\begin{center}\includegraphics[width=0.8\textwidth]{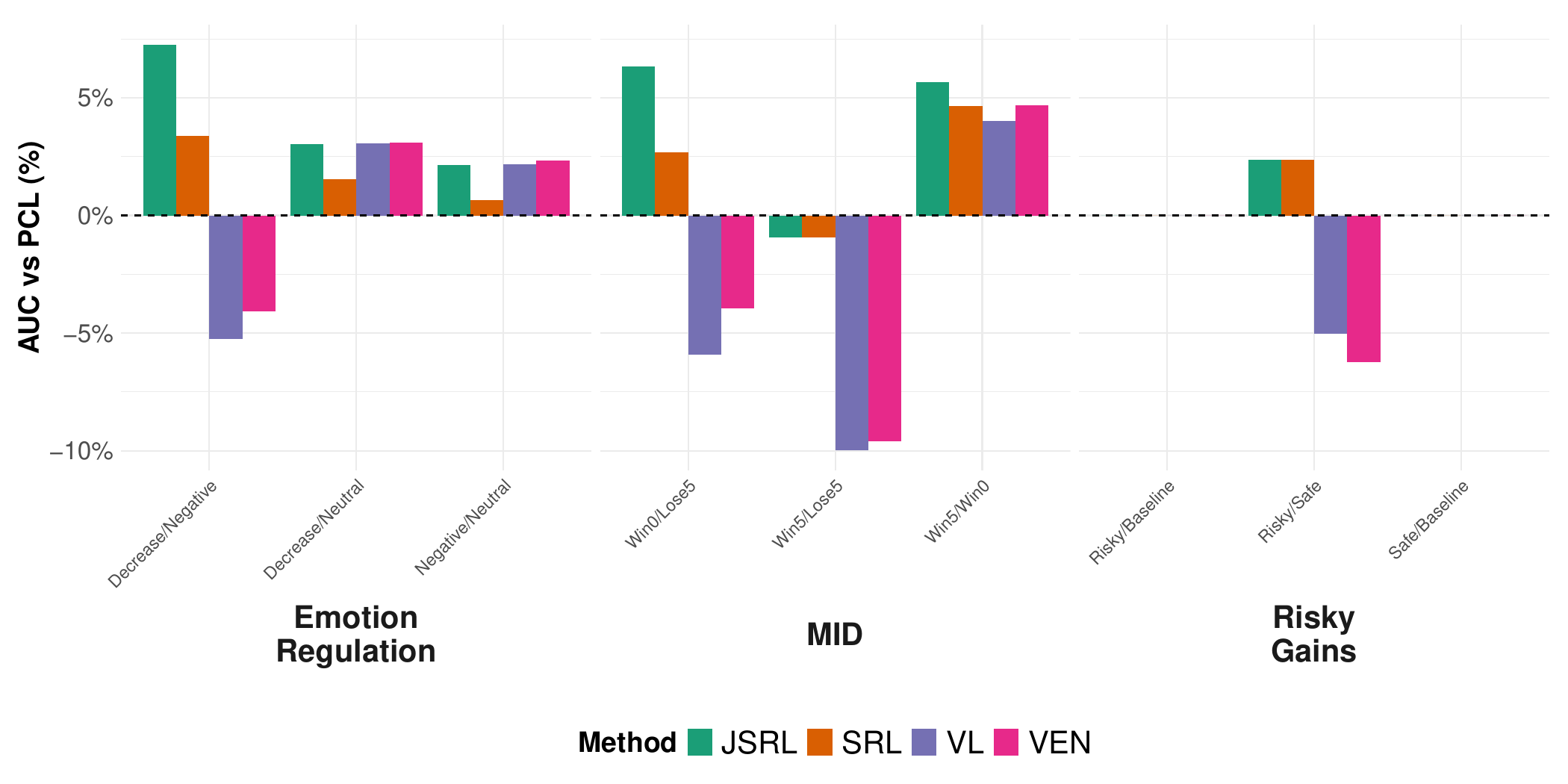}
\caption{All models AUC is reported as its relative percentage improvement over the Principal Component LASSO (PCL) model. Note that no data is displayed for the Risky/Baseline or Safe/Baseline tasks for Risky Gains as all models achieved an AUC of 1.}
\label{figs1}\end{center}\end{figure}

\begin{figure}[H]\begin{center}\includegraphics[width=0.8\textwidth]{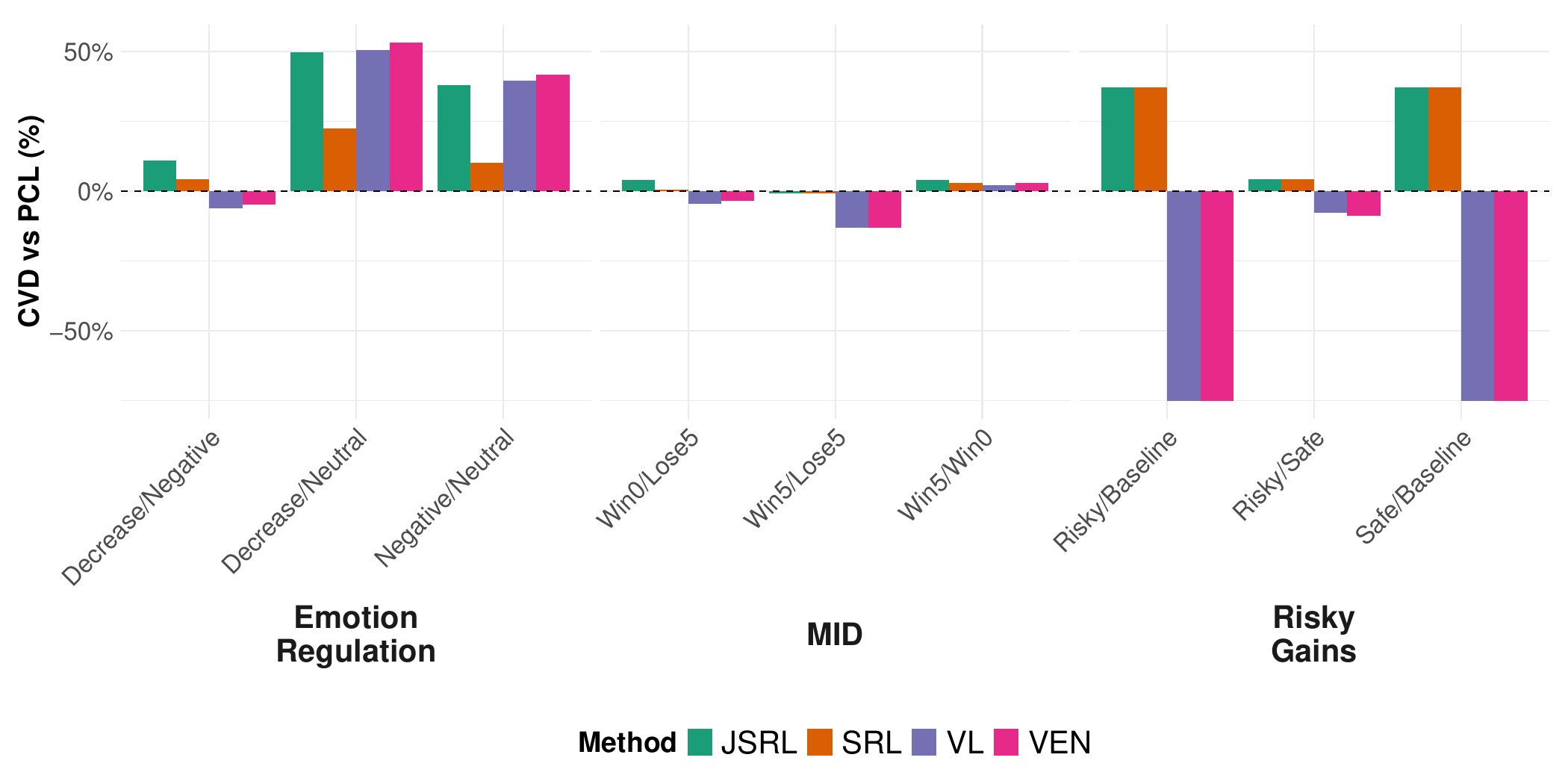}
\caption{All models CVD is reported as its relative percentage improvement over the Principal Component LASSO (PCL) model. Negative changes for voxel models in the Risky Gains dataset are clipped at -75\% to retain the scale of the figure.}
\label{figs2}\end{center}\end{figure}

\begin{figure}[H]\begin{center}\includegraphics[width=0.8\textwidth]{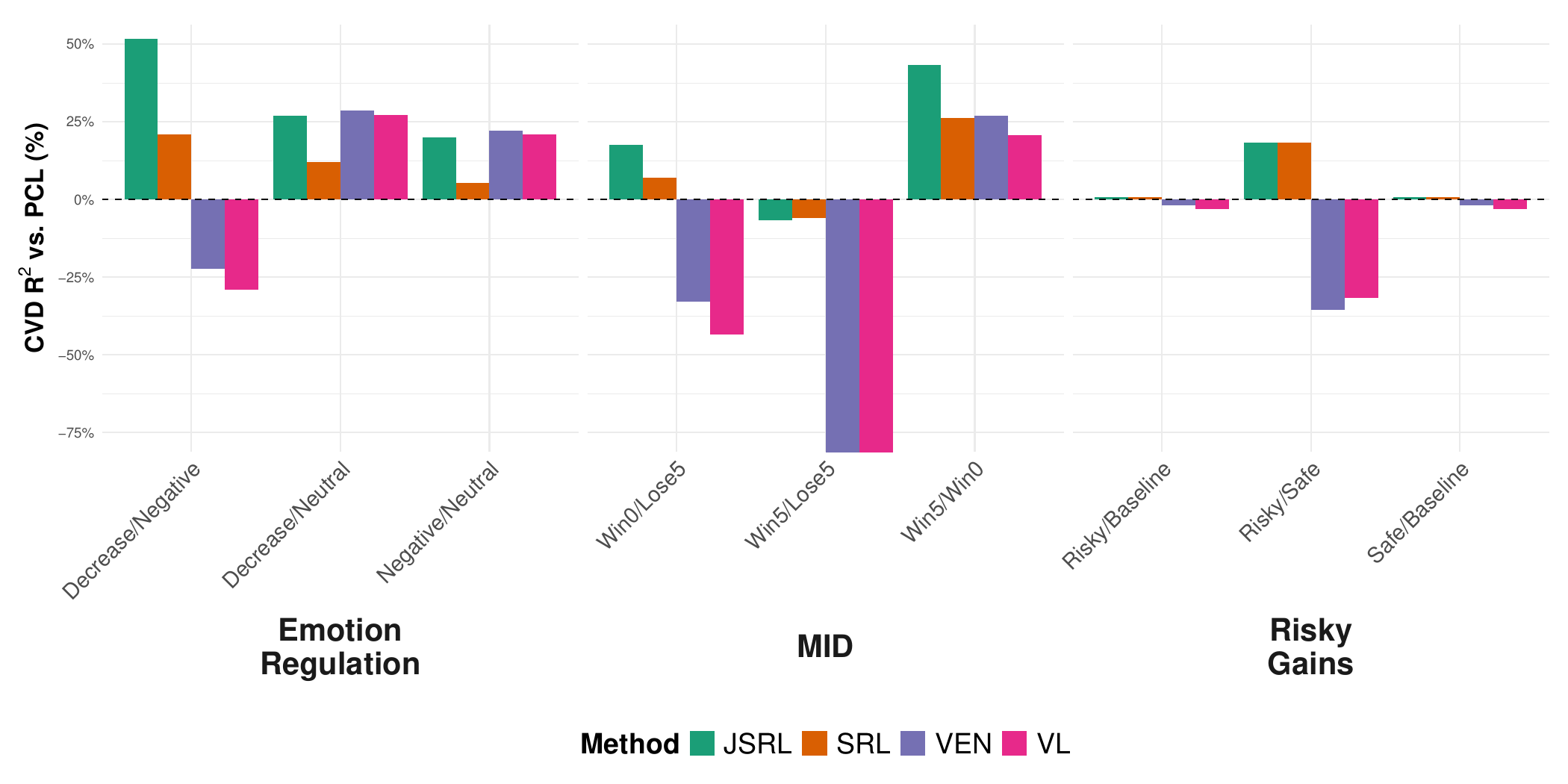}
\caption{All models CVD $R^2$ is reported as its relative percentage improvement over the Principal Component LASSO (PCL) model. Negative changes for voxel models in the MID dataset are clipped at -75\% to retain the scale of the figure.}
\label{figs3}\end{center}\end{figure}

\printbibliography

\appendix

\end{document}